\documentclass{article}

\usepackage{times}
\usepackage{graphicx} 
\usepackage{subcaption} 
\usepackage{natbib}
\usepackage{algorithm}
\usepackage{algorithmic}
\usepackage{hyperref}

\usepackage{url}
\usepackage{mathtools}
\usepackage{amsmath}
\usepackage{amsfonts}
\usepackage{amsthm}

\usepackage[accepted]{icml2016} 

\newcommand{\half}{\frac{1}{2}} 

\newcommand{\ra}{\rangle}  
\newcommand{\la}{\langle}  
\newcommand{\st}{\hbox{ \,\,subject to\,\, }}  
\newcommand{\matb}{\left( \begin{matrix*}[r] }  
\newcommand{\mate}{\end{matrix*}\right)}

\newcommand{\eqb}[1]{\begin{equation}\label{#1}}
\newcommand{\eqe}{\end{equation}}

\newcommand{\aln}[1]{\begin{align}#1\end{align}}
\newcommand{\splt}[1]{\begin{split}#1\end{split}}
\newcommand{\css}[1]{\begin{cases}#1\end{cases}}

\newcommand{\lm}{_{l-1}}

\DeclareMathOperator*{\minimize}{minimize\quad}
\DeclareMathOperator*{\subjectto}{subject\,to\quad}

\DeclareMathOperator*{\argmin}{arg\,min}

\usepackage{color}

\graphicspath{{./graphs/}}

\icmltitlerunning{Scaling Neural Networks with ADMM}

\begin{document} 
\twocolumn[
\icmltitle{Training Neural Networks Without Gradients:  \\ A Scalable ADMM Approach}

\icmlauthor{Gavin Taylor$^1$}{taylor@usna.edu}
\icmlauthor{Ryan Burmeister$^1$}{}
\icmlauthor{Zheng Xu$^2$}{xuzh@cs.umd.edu}
\icmlauthor{Bharat Singh$^2$}{bharat@cs.umd.edu}
\icmlauthor{Ankit Patel$^3$}{abp4@rice.edu}
\icmlauthor{Tom Goldstein$^2$}{tomg@cs.umd.edu}
\icmladdress{$^1$United States Naval Academy, Annapolis, MD USA}
\vspace{-.3cm}\icmladdress{$^2$University of Maryland, College Park, MD USA}
\vspace{-.3cm}\icmladdress{$^3$Rice University, Houston, TX USA}

\icmlkeywords{boring formatting information, machine learning, ICML}
\vspace{-3mm}
\vskip 0.3in
]

\begin{abstract} 
With the growing importance of large network models and enormous training
datasets, GPUs have become increasingly necessary to train neural networks.
This is largely because conventional optimization algorithms rely on
stochastic gradient methods that don't scale well to large numbers of cores in
a cluster setting.  Furthermore, the convergence of all gradient methods,
including batch methods,  suffers from common problems like saturation
effects, poor conditioning, and saddle points.  This paper explores an
unconventional training method that uses alternating direction methods and
Bregman iteration to train networks without gradient descent steps.  The
proposed method reduces the network training problem to a sequence of
minimization sub-steps that can each be solved {\em globally} in closed form.
The proposed method is advantageous because it avoids many of the caveats that
make gradient methods slow on highly non-convex problems.  The
method exhibits strong scaling in the distributed setting, yielding linear
speedups even when split over thousands of cores. 

\end{abstract} 

\section{Introduction}

As hardware and algorithms advance, neural network performance is constantly
improving for many machine learning tasks. This is particularly true in
applications where extremely large datasets are available to train models with
many parameters.  Because big datasets provide results that (often
dramatically) outperform the prior state-of-the-art in many machine learning
tasks, researchers are willing to purchase specialized hardware such as GPUs,
and commit large amounts of time to training models and tuning
hyper-parameters.

Gradient-based training methods have several properties that contribute to this need for specialized hardware.  First, while large amounts of data
can be shared amongst many cores, existing optimization methods suffer when
parallelized.
 Second, training neural nets requires optimizing highly non-convex objectives
 that exhibit saddle points, poor conditioning, and vanishing gradients, all
 of which slow down gradient-based methods such as stochastic gradient descent,
conjugate gradients, and BFGS.  Several mitigating approaches to avoiding this issue have been
introduced, including rectified linear units (ReLU)~\cite{nair2010rectified},
Long Short-Term Memory networks~\cite{hochreiter1997long},  RPROP \cite{riedmiller1993direct}, and others, but the
fundamental problem remains.

In this paper, we introduce a new method for training the parameters of neural
nets using the Alternating Direction Method of Multipliers (ADMM) and Bregman
iteration.  This approach addresses several problems facing classical gradient
methods; the proposed method exhibits linear scaling when data is parallelized
across cores, and is robust to gradient saturation and poor conditioning.  The
method decomposes network training into a sequence of sub-steps that are each
solved to global optimality.  The scalability of the proposed method, combined
with the ability to avoid local minima by globally solving each substep, can
lead to dramatic speedups.

We begin in Section \ref{sec:background} by describing the mathematical
notation and context, and providing a discussion of several weaknesses of
gradient-based methods that we hope to address.  Sections \ref{sec:min} and
\ref{sec:lagrange} introduce and describe the optimization approach, and
Sections \ref{sec:dist} and \ref{sec:implementation} describe in detail the
distributed implementation.  Section \ref{sec:experiments} provides an
experimental comparison of the new approach with standard implementations of
several gradient-based methods on two problems of differing size and
difficulty.  Finally, Section \ref{sec:discussion} contains a closing
discussion of the paper's contributions and the future work needed.

\section{Background and notation}
\label{sec:background}

Though there are many variations, a typical neural network consists of $L$
layers, each of which is defined by a linear operator $W_l,$ and a non-linear
neural activation function $h_l.$  Given a (column) vector of input
activations, $a\lm,$ a single layer computes and outputs the non-linear
function $a_l=h_l(W_la_{l-1}).$  A network is formed by layering these units
together in a nested fashion to compute a composite function; in the
3-layer case, for example, this would be
     \aln{ \label{general}
      f(a_0; W) =  W_3( h_2(W_2 h_1(W_1a_0)))
      }
where $W=\{W_l\}$ denotes the ensemble of weight matrices, and $a_0$ contains
input activations for every training sample (one sample per column).  The
function $h_3$ is absent as it is common for the last layer to not have an
activation function.

Training the network is the task of tuning the weight matrices $W$ to match
the output activations $a_L$ to the targets $y$, given the inputs $a_0.$   Using a loss function
$\ell,$ the training problem can be posed as
\aln{ \label{originalMin}
\min_W \,\,\,   \ell( f(a_0; W), y  ) 
   }
Note that in our notation, we have included all input activations for all
training data into the matrix/tensor $a_0.$  This notation benefits our
discussion of the proposed algorithm, which operates on all training data
simultaneously as a batch.

Also, in our formulation the tensor $W$ contains {\em linear operators}, but
not necessarily dense matrices.  These linear operators can be convolutions
with an ensemble of filters, in which case \eqref{general} represents a
convolutional net.
 
Finally, the formulation used here assumes a feed-forward architecture.
However, our proposed methods can handle more complex network topologies (such
as recurrent networks) with little modification.  

\subsection{What's wrong with backprop?}

Most networks are trained using stochastic gradient descent (SGD, i.e.
backpropagation) in which the gradient of the network loss function is
approximated using a small number of training samples, and then a descent step
is taken using this approximate gradient.  Stochastic gradient methods work
extremely well in the serial setting, but lack scalability.  Recent attempts
to scale SGD include Downpour, which runs SGD simultaneously on multiple
cores.  This model averages parameters across cores using multiple
communication nodes that store copies of the model. A conceptually similar
approach is elastic averaging \cite{zhang2015deep}, in which different
processors simultaneously run SGD using a quadratic penalty term that prevents
different processes from drifting too far from the central average.   These
methods have found success with modest numbers of processors, but fail to
maintain strong scaling for large numbers of cores.  For example, for several
experiments reported in~\cite{dean12}, the Downpour distributed SGD method
runs slower with 1500 cores than with 500 cores.

The scalability of SGD is limited because it relies on a {\em large} number of
inexpensive minimization steps that each use a {\em small} amount of data.
Forming a noisy gradient from a small mini-batch requires very little
computation.  The low cost of this step is an asset in the serial setting
where it enables the algorithm to move quickly, but disadvantageous in the
parallel setting where each step is too inexpensive to be split over multiple
processors.  For this reason, SGD is ideally suited for computation on GPUs,
where multiple cores can simultaneously work on a small batch of data using a
shared memory space with virtually no communication overhead.   

When parallelizing over CPUs, it is preferable to have methods that use a {\em
small} number of {\em expensive} minimization steps, preferably involving a
large number of data.  The work required on each minimization step can then be
split across many worker nodes, and the latency of communication is amortized
over a large amount of computation.  This approach has been suggested by
numerous authors who propose batch computation methods
\cite{ngiam2011optimization}, which compute exact gradients on each iteration
using the entire dataset, including conjugate gradients
\cite{towsey1995training,moller1993scaled}, BFGS, and Hessian-free
\cite{martens2011learning,sainath2013accelerating} methods.

Unfortunately, all gradient-based approaches, whether batched or stochastic,
also suffer from several other critical drawbacks.  First, gradient-based methods suffer from the {\em vanishing gradients.}
  During backpropagation, the derivative of shallow layers in a network are
  formed using products of weight matrices and derivatives of non-linearities
  from downstream layers.  When the eigenvalues of the weight matrices are
  small and the derivatives of non-linearities are nearly zero (as they often
  are for sigmoid and ReLU non-linearities), multiplication by these terms
  annihilates information. The resulting gradients in shallow layers contain
  little information about the error
  \cite{bengio1994learning,riedmiller1993direct,hochreiter1997long}.

Second, backprop has the potential to get stuck at local minima and saddle
  points.  While recent results suggest that local minimizers of SGD are close
  to global minima \cite{choromanska2014loss}, in practice SGD often lingers
  near saddle points where gradients are small \cite{dauphin2014identifying}.  

Finally, backprop does not easily parallelize over layers, a significant
  bottleneck when considering deep architectures.  However, recent work on SGD
  has successfully used model parallelism by using multiple replicas of the
  entire network \cite{dean12}.  

We propose a solution that helps alleviate these problems by separating the
objective function at each layer of a neural network into two terms: one term
measuring the relation between the weights and the input activations, and the
other term containing the nonlinear activation function. We then apply an
alternating direction method that addresses each term separately.  The first
term allows the weights to be updated without the effects of vanishing
gradients. In the second step, we have a non-convex minimization problem that
can be solved \emph{globally} in closed-form. Also, the form of the
objective allows the weights of every layer to be updated
independently, enabling parallelization over layers.

This approach does not require any gradient steps at all.  Rather, the problem
of training network parameters is reduced to a series of minimization
sub-problems using the alternating direction methods of multipliers.  These
minimization sub-problems are solved globally in closed form. 

\subsection{Related work}
Other works have applied least-squares based methods to neural networks.  One notable example is the method of auxiliary coordinates (MAC) \cite{carreira2012distributed} which uses quadratic penalties to approximately enforce equality constraints.  Unlike our method, MAC requires iterative solvers for sub-problems, whereas the method proposed here is designed so that all sub-problems have closed form solutions. Also unlike MAC, the method proposed here uses Lagrange multipliers to exactly enforce equality constraints, which we have found to be necessary for training deeper networks.  

Another related approach is the expectation-maximization (EM) algorithm of \cite{patel2015drm}, which is derived from the Deep Rendering Model (DRM), a hierarchical generative model for natural images. They show that feedforward propagation in a deep convolutional net corresponds to inference on their proposed DRM. They derive a new EM learning algorithm for their proposed DRM that employs least-squares parameter updates that are conceptually similar to (but different from) the Parallel Weight Update proposed here (see Section~\ref{sec:dist}). However, there is currently no implementation nor any training results to compare against.

Note that our work is the first to consider alternating least squares as a method to distribute computation across a cluster, although the authors of \cite{carreira2012distributed} do consider implementations that are ``distributed'' in the sense of using multiple threads on a single machine via the Matlab matrix toolbox.  

\section{Alternating minimization for neural networks}
\label{sec:min}
The idea behind our method is to decouple the weights from the nonlinear link
functions using a splitting technique.  Rather than feeding the output of the
linear operator $W_l$ directly into the activation function $h_l,$ we store the
output of layer $l$ in a new variable $z_l = W_la_{l-1}.$  We also represent
the output of the link function as  a vector of activations $a_l=h_l(z_l).$
We then wish to solve the following problem
\aln{ \splt{ \label{constrained}
    \minimize_{\hspace{-3mm}\{W_l\},\{a_l\},\{z_l\}}\,\,& \ell(z_L, y) \\
    \subjectto & z_l = W_la_{l-1},  \text{ for } l=1,2,\cdots L\\
  &  a_l=h_l(z_l) ,\text{ for } l=1,2,\cdots L-1.
  }}
Observe that solving \eqref{constrained} is equivalent to solving
\eqref{originalMin}.  Rather than try to solve \eqref{constrained} directly,
we relax the constraints by adding an $\ell_2$ penalty function to the
objective and attack the unconstrained problem
\aln{  \label{split}
&\minimize_{\hspace{-3mm}\{W_l\},\{a_l\},\{z_l\}}\,\, \ell(z_L, y) +
  \beta_{L}\|z_{L} -  W_{L}a_{L-1} \|^2  \nonumber\\  & \qquad+
  \sum_{l=1}^{L-1}
  \left[\gamma_l\| a_l-h_l(z_l) \|^2 +  \beta_l\|z_l -  W_la_{l-1} \|^2\right]
  }
where $\{\gamma_l\}$ and $\{\beta_l\}$ are constants that control the weight
of each constraint.  The formulation \eqref{split} only {\em approximately}
enforces the constraints in \eqref{constrained}.  To obtain exact enforcement
of the constraints, we add a Lagrange multiplier term to \eqref{split}, which yields
\aln{  \label{bregman}
& \hspace{-3mm} \minimize_{\hspace{-1mm}\{W_l\},\{a_l\},\{z_l\}}\,\, \ell(z_L,
  y)    \\
 &  \hspace{7mm} + \la z_{L},\lambda \ra+  \beta_{L}\|z_{L} -  W_{L}a_{L-1} \|^2 \nonumber \\
 &\hspace{7mm} + \sum_{l=1}^{L-1} \left[\gamma_l\| a_l-h_l(z_l) \|^2 +  \beta_l\|z_l
-  W_la_{l-1} \|^2\right]. \nonumber
  }
where $\lambda$ is a vector of Lagrange multipliers with the same dimensions
as $z_{L}.$  Note that in a classical ADMM formulation, a Lagrange
multiplier would be added for each constraint in \eqref{constrained}.  The
formulation above corresponds more closely to Bregman iteration, which only
requires a Lagrange correction to be added to the objective term (and not the
constraint terms), rather than classical ADMM.  We have found the Bregman
formulation to be far more stable than a full scale ADMM formulation.  This
issue will be discussed in detail in Section \ref{sec:lagrange}.
 
The split formulation \eqref{split} is carefully designed to be easily
minimized using an alternating direction method in which each sub-step has
a simple closed-form solution.  The alternating direction scheme proceeds by
updating one set of variables at a time -- either $\{W_l\},$ $\{a_l\},$ or
$\{z_l\}$ -- while holding the others constant.  The simplicity of the
proposed scheme comes from the following observation:  The minimization of
\eqref{split} with respect to both $\{W_l\}$ and $\{a_{l-1}\}$ is a simple
linear least-squares problem.   
Only the minimization of \eqref{split} with respect $\{z_l\}$ is nonlinear.
However, there is no coupling between the entries of  $\{z_l\},$ and so the
problem of minimizing for $\{z_l\}$ decomposes into solving a large number of
one-dimensional problems, one for each entry in $\{z_l\}.$  Because each
sub-problem has a simple form and only 1 variable, these problems can be solved
{\em globally} in closed form.
   
The full alternating direction method is listed in Algorithm~\ref{alg}. We discuss the details below.

\begin{algorithm}[tb]
   \caption{ADMM for Neural Nets}
   \label{alg}
\begin{algorithmic}
   \STATE {\bfseries Input:} training features $\{a_0\},$ and labels $\{y\},$ 
   \STATE {\bfseries Initialize:} allocate $\{a_l\}_{l=1}^{L=1},$  $\{z_l\}_{l=1}^L,$ and $\lambda$
   \STATE\textbf{repeat}
   \FOR{$l=1,2,\cdots,L-1$}
    \STATE $W_l \gets  z_l a_{l-1}^\dagger$
   \STATE  {\small $a_l \hspace{-1mm}\gets \hspace{-1mm}(\beta_{l+1} W_{l+1}^TW_{l+1}+\gamma_{l} I)^{-1} (\beta_{l+1}
  W_{l+1}^Tz_{l+1} +\gamma_l h_l (z_l))$}
  \STATE $z_l\gets\argmin_{z}\gamma_l\| a_l-h_l(z) \|^2 +  \beta_l\|z_l -  W_la_{l-1} \|^2$
    \ENDFOR
     \STATE $W_L \gets  z_L a_{L-1}^\dagger$ 
     \STATE $z_L\gets\argmin_{z} \ell(z,y) + \la z_{L},\lambda \ra+  \beta_L\|z -  W_La_{l-1} \|^2$

  \STATE $ \lambda \gets \lambda+ \beta_{L}(z_{L} -  W_{L}a_{L-1})$
   \STATE{\textbf{until} converged}
   \end{algorithmic}
\end{algorithm}

\subsection{Minimization sub-steps} 
In this section, we consider the updates for each variable in \eqref{bregman}.  The algorithm proceeds by minimizing for $W_l,$ $a_l,$ and $z_l,$ and then updating the Lagrange multipliers $\lambda.$

\paragraph{\textbf{Weight update }} We first consider the minimization of
\eqref{split} with respect to $\{W_l\}.$
For each layer $l,$ the optimal solution minimizes $\|z_l -  W_la_{l-1} \|^2.$
This is simply a least squares problem, and the solution is given by
$W_l \gets  z_l a_{l-1}^\dagger$
where  $a_{l-1}^\dagger$ represents the pseudoinverse of the (rectangular)
activation matrix $a_{l-1}.$

\paragraph{\textbf{Activations update }}
Minimization for $a_l$ is a simple least-squares problem similar to the weight update.  However, in this case the matrix $a_l$ appears in two penalty terms in \eqref{split}, and so we must minimize 
$\beta_l\|z_{l+1}-W_{l+1}a_l\|+\gamma_l\| a_l-h_l(z_l) \|^2$ 
for $a_l,$ holding all other variables fixed.  The new value of  $a_l$ is given by
\aln{\label{acts_update} 
  \hspace{-3mm} (\beta_{l+1} W_{l+1}^TW_{l+1}+\gamma_{l} I)^{-1} (\beta_{l+1}
  W_{l+1}^Tz_{l+1} +\gamma_l h_l (z_l))
 }
 where $W_{l+1}^T$ is the adjoint (transpose) of $W_{l+1}.$ 

\paragraph{\textbf{Outputs update }} The update for $z_l$ requires minimizing
\aln{\label{outs_update}
  \min_{z}\gamma_l\| a_l-h_l(z) \|^2 +  \beta_l\|z -  W_la_{l-1} \|^2.
}
This problem is non-convex and non-quadratic (because of the non-linear term
$h$).  Fortunately, because the non-linearity $h$ works entry-wise on its
argument, the entries in  $z_l$ are de-coupled.
Solving \eqref{outs_update} is particularly easy when $h$ is piecewise linear,
as it can be solved in closed form; common piecewise linear choices for $h$
include rectified linear units (ReLUs) and non-differentiable sigmoid
functions given by
$$
h_{relu}(x)=\begin{cases}
x, \text{ if }  x>0\\
0, \text{ otherwise}
\end{cases} \hspace{-4mm},
 h_{sig}(x)=\begin{cases}
1, \text{ if }  x \ge 1\\
x, \text{ if }  0<x<1\\
0, \text{ otherwise}
\end{cases} \hspace{-4mm}. 
$$
For such choices of $h,$ the minimizer of \eqref{outs_update} is easily
computed using simple if-then logic.  For more sophistical choices of $h,$
including smooth sigmoid curves, the problem can be solved quickly with a
lookup table of pre-computed solutions because each 1-dimensional problem only
depends on two inputs.

\paragraph{\textbf{Lagrange multiplier update }}
After minimizing for $\{W_l\},$ $\{a_l\},$ and $\{z_l\},$ the Lagrange
multiplier update is given simply by
 \aln{\label{lagrange}
 \lambda \gets \lambda+ \beta_{L}(z_{L} -  W_{L}a_{L-1}).
 }
We discuss this update further in Section \ref{sec:lagrange}.


\section{Lagrange multiplier updates via method of multipliers and Bregman iteration}
\label{sec:lagrange}

The proposed method can be viewed as solving the constrained problem
\eqref{constrained} using Bregman iteration, which is closely related to ADMM.
The convergence of Bregman iteration is fairly well understood in the presence
of linear constraints \cite{yin2008bregman}.   The convergence of ADMM is
fairly well understood for convex problems involving only two separate
variable blocks \cite{he2015non}.   Convergence results also guarantee that a
local minima is obtained for two-block non-convex objectives under certain
smoothness assumptions \cite{nocedal2006numerical}.  

Because the proposed scheme involves more than two coupled variable blocks and
a non-smooth penalty function, it lies outside the scope of known convergence
results for ADMM.  In fact, when ADMM is applied to \eqref{constrained} in a
conventional way using separate Lagrange multiplier vectors for each
constraint, the method is highly unstable because of the de-stabilizing effect
of a large number of coupled, non-smooth, non-convex terms. 

Fortunately, we will see below that the Bregman Lagrange update method
\eqref{lagrange} does not involve any non-smooth constraint terms, and
the resulting method seems to be extremely stable.

\subsection{Bregman interpretation}

Bregman iteration (also known as the method of multipliers) is a general framework for solving constrained optimization problems.  Methods of this type have been used extensively in the sparse optimization literature \cite{yin2008bregman}. Consider the general problem of minimizing 
\aln{ \label{simple_constrained}
\min_u J(u)  \st Au=b
}
for some convex function $J$ and linear operator $A.$
Bregman iteration repeatedly solves
\aln{\label{breg_iter}
 u^{k+1} \gets \min D_J(u,u^k) + \half \|Au-b\|^2 
 }
where $p\in \partial J (u^k)$ is a (sub-)gradient of $J$ at $u^k,$ and  $D_J(u,u^k)  = J(u)-J(u^k) - \la u-u^k, p\ra $ is the so-called Bregman distance. The iterative process \eqref{breg_iter} can be viewed as minimizing the objective $J$ subject to an inexact penalty that approximately obtains $Ax\approx b,$ and then adding a linear term to the objective to weaken it so that the quadratic penalty becomes more influential on the next iteration.  

The Lagrange update described in Section \ref{sec:min} can be interpreted as
performing Bregman iteration to solve the problem \eqref{constrained}, where
$J(u) = \ell(z_L, y),$ and $A$ contains the constraints in
\eqref{constrained}.  On each iteration, the outputs $z_{l}$ are updated
immediately before the Lagrange step is taken, and so $z\lm$ satisfies the
optimality condition
  $$0\in \partial_z  \ell(z_{L}, y)+  \beta_{L}(z_{L} -  W_{L}a_{L-1} )+  \lambda.$$ 
  It follows that
  $$  \lambda + \beta_{L}(z_{L} -  W_{L}a_{L-1} ) \in -\partial_z  \ell(z_{L},
  y).$$
For this reason, the Lagrange update \eqref{lagrange} can be interpreted as updating the sub-gradient in the Bregman iterative method for solving \eqref{constrained}.  The combination of the Bregman iterative update with an alternating minimization strategy makes the proposed algorithm an instance of the split Bregman method \cite{goldstein2009split}.

\subsection{Interpretation as method of multipliers}
In addition to the Bregman interpretation, the proposed method can also be viewed as an approximation to the method of multipliers, which solves constrained problems of the form
\aln{ \label{simple_constrained}
\min_u J(u)  \st Au=b
}
for some convex function $J$ and (possibly non-linear) operator $A.$  In its most general form (which does not assume linear constraints) the method proceeds using the iterative updates
$$\css{ 
 u^{k+1} & \gets \min  J(u) + \la \lambda^k, A(u)-b\ra  +\frac{\beta}{2} \|A(u)-b\|^2  \\
  \lambda^{k+1}& \gets  \lambda^k+\partial_u\{\frac{\beta}{2} \|A(u)-b\|^2\}
  }$$
where $\lambda^k$ is a vector of Lagrange multipliers that is generally initialized to zero, and $\frac{\beta}{2} \|A(u)-b\|^2$ is a quadratic penalty term.  After each minimization sub-problem, the gradient of the penalty term is added to the Lagrange multipliers.  When the operator $A$ is linear, this update takes the form  $\lambda^{k+1} \gets  \lambda^k+\beta A^T (Au-b),$ which is the most common form of the method of multipliers. 

Just like in the Bregman case, we now let $J(u) = \ell(z_{L}, y),$ and let
$A$ contain the constraints in \eqref{constrained}.  After a minimization
pass, we must update the Lagrange multiplier vector.  Assuming a good
minimizer has been achieved, the derivative of \eqref{bregman} should be
nearly zero.  All variables {\em except} $z_{L}$ appear only in the
quadratic penalty, and so these derivatives should be negligibly small.  The
only major contributor to the gradient of the penalty term is
$z_{L},$ which appears in both the loss function and the quadratic penalty.
The gradient of the penalty term with respect to $z_{L},$ is $ \beta_{L}(z_{L}
-  W_{L}a_{L-1} ),$ which is exactly the proposed multiplier update.

When the objective is approximately minimized by alternately updating separate
blocks of variables (as in the proposed method), this becomes an instance of
the ADMM~\cite{boyd2011distributed}. 

\section{Distributed implementation using data parallelism} \label{sec:dist}
The main advantage of the proposed alternating minimization method is its high
degree of scalability.  In this section, we explain how the method is
distributed.

Consider distributing the algorithm across $N$ worker nodes.  The ADMM method
is scaled using a data parallelization strategy, in which different nodes
store activations and outputs corresponding to different subsets of the
training data.  For each layer, the activation matrix is broken into columns
subsets as $a_i = (a_1,a_2,\cdots, a_N).$  The output matrix $z_l$ and
Lagrange multipliers $\lambda$ decompose similarly.   
  
The optimization sub-steps for updating $\{a_l\}$ and $\{z_l\}$ do not require
any communication and parallelize trivially.  The weight matrix update
requires the computation of pseudo-inverses and products involving the
matrices $\{a_l\}$ and $\{z_l\}.$  This can be done effectively using {\em
transpose reduction}  strategies that reduce the dimensionality of matrices
before they are transmitted to a central node.
    
\paragraph{\textbf{Parallel Weight update }} The weight update has the form $W_l \gets  z_l a_l^\dagger,$ where  $a_l^\dagger$ represents the pseudoinverse of the activation matrix $a_l.$  This pseudoinverse can be written $a_l^\dagger = a_l^T(a_la_l^T)^{-1}.$  Using this expansion, the W update decomposes across nodes as   
      $$W_l \gets \left( \sum_{n=1}^N  z_l^n (a_l^n)^T  \right)  \left( \sum_{n=1}^N a_l^n(a_l^n)^T \right)^{-1}.$$
The individual products $z_l^n (a_l^n)^T$ and $a_l^n(a_l^n)^T$ are computed
separately on each node, and then summed across nodes using a single reduce
operation.  Note that the width of $a_l^n$ equals the number of training
vectors that are stored on node $n,$ which is potentially very large for big
data sets.  When the number of features (the number of rows in $a_l^n$) is
less than the number of training data (columns of $a_l^n$), we can exploit
transpose reduction when forming these products -- the product
$a_l^n(a_l^n)^T$ is much smaller than the matrix $a_l^n$ alone.  This
dramatically reduces the quantity of data transmitted during the reduce
operation.   
 
   Once these products have been formed and reduced onto a central server, the central node computes the inverse of  $a_la_l^T,$ updates $W_l,$ and then broadcasts the result to the worker nodes. 

\paragraph{\textbf{Parallel Activations update }}  The update \eqref{acts_update} trivially decomposes across workers, with each worker computing
$$  a_l^n \hspace{-.5mm}\gets  \hspace{-.5mm}(\beta_{l+1} W_{l+1}^TW_{l+1}+\gamma I)^{-1}(\beta_{l+1}
W_{l+1}^Tz_{l+1}^n +\gamma_l h_l (z_l^n)).$$
Each server maintains a full representation of the entire weight matrix, and
can formulate its own local copy of the matrix inverse $(\beta_{l+1}
W_{l+1}^TW_{l+1}+\gamma I)^{-1}.$

\paragraph{\textbf{Parallel Outputs update }} Like the activations update, the update for $z_l$ trivially parallelizes and each worker node solves
\aln{\label{par_outs_update}
  \min_{z_l^n}\gamma_l\| a_l^n-h_l(z_l^n) \|^2 +  \beta_l\|z_l^n -  W_la_{l-1}^n \|^2.
}
Each worker node simply computes $W_la_{l-1}^n$ using local data, and then
updates each of the (decoupled) entries in $z_l^n$ by solving a 1-dimensional
problem in closed form. 

\paragraph{\textbf{Parallel Lagrange multiplier update }}
The Lagrange multiplier update also trivially splits across nodes, with worker $n$ computing
 \aln{\label{lagrange}
 \lambda^n \gets \lambda^n+ \beta_{L}(z_{L}^n -  W_{L}a_{L-1}^n)
 }
 using only local data.

\section{Implementation details}
\label{sec:implementation}
Like many training methods for neural networks, the ADMM approach requires
several tips and tricks to get maximum performance. The convergence theory for
the method of multipliers requires a good minimizer to be computed before
updating the Lagrange multipliers.  When the method is initialized with random
starting values, the initial iterates are generally far from optimal.  For
this reason, we frequently ``warm start'' the ADMM method by running several
iterations without Lagrange multiplier updates.  

The method potentially requires the user to choose a large number of
parameters $\{\gamma_i\}$ and $\{\beta_i\}.$  We choose $\gamma_i=10$ and
$\beta_i=1$ for all trials runs reported here, and we have found that this
choice works reliably for a wide range of problems and network architectures.
Note that in the classical ADMM method, convergence is guaranteed for any
choice of the quadratic penalty parameters.

We use training data with binary class labels, in which each output entry
$a_L$ is either 1 or 0.  We use a separable loss function with a hinge
penalty of the form 
$$\ell(z,a)= \css{
\max\{1-z,0\},\text{ when } a=1, \\
\max\{a,0\},\text{ when } a=0.
}$$
This loss function works well in practice, and yields minimization sub-problems that are easily solved in closed form.

Finally, our implementation simply initializes the activation matrices
$\{a_l\}$ and output matrices $\{z_l\}$ using i.i.d Gaussian random variables.
Because our method updates the weights before anything else, the weight
matrices do not require any initialization.  The results presented here are
using Gaussian random variables with unit variance, and the results seem to be
fairly insensitive to the variance of this distribution.  This seems to be
because the output updates are solved to global optimality on each iteration.

\section{Experiments}
\label{sec:experiments}

\begin{figure*}[t]
\vspace{-3mm}
  \centering
  \hfill
  \begin{subfigure}[t]{0.45\textwidth}
    \includegraphics[width=\textwidth]{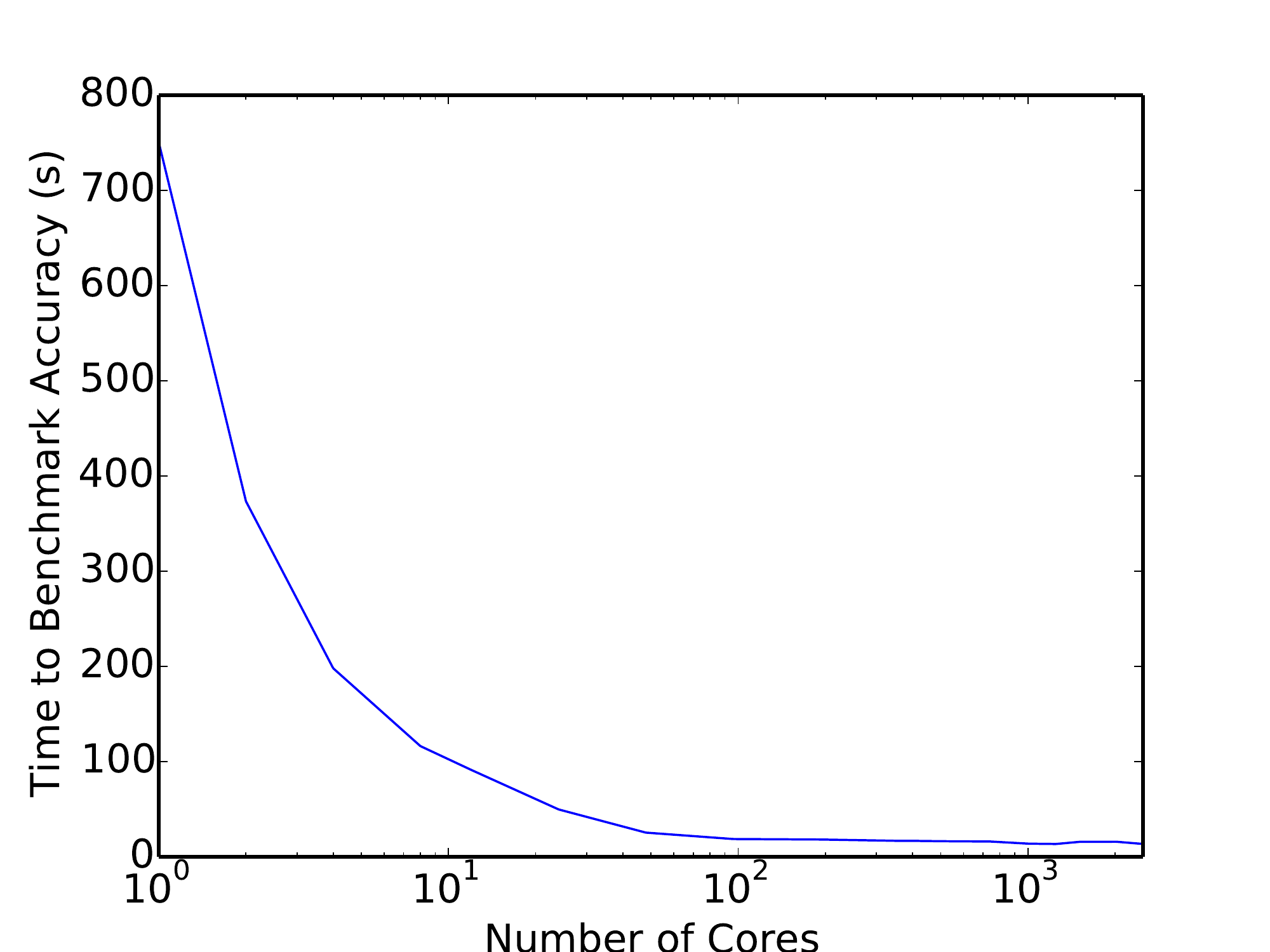}
    \caption{\textbf{Time required for ADMM to reach 95\% test accuracy vs number of cores}.  This problem was not large
  enough to support parallelization over many cores, yet the advantages of
  scaling are still apparent (note the x-axis has log scale).
In comparison, on the GPU, L-BFGS reached this threshold in 3.2
seconds, CG in 9.3 seconds, and SGD
in 8.2 seconds.}
    \label{sfig:svScaling}
  \end{subfigure}
  \hfill
  \begin{subfigure}[t]{0.45\textwidth}
    \includegraphics[width=\textwidth]{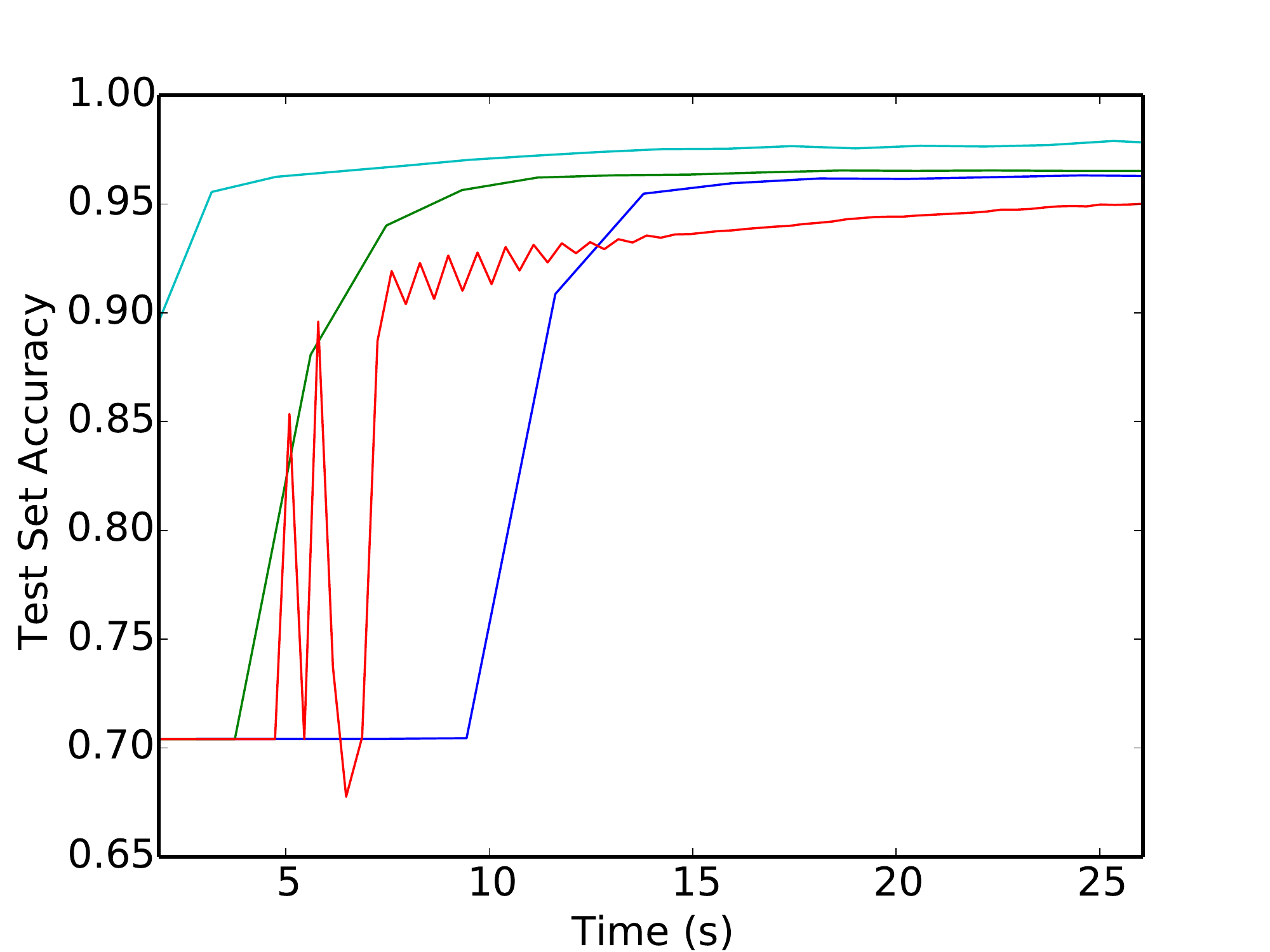}
    \caption{\textbf{Test set predictive accuracy as a function of time in
      seconds} for ADMM on 2,496 cores (blue), in addition to GPU
    implementations of conjugate gradients (green), SGD (red), and L-BFGS
  (cyan).}
    \label{sfig:svConverge}
  \end{subfigure}
  \hfill
  \caption{
  Street View House Numbers (subsection \ref{ssec:svhn})
  \vspace{-3mm}}
  \label{fig:scaling}
\end{figure*}

\begin{figure*}[t]
\vspace{-3mm}
  \centering
  \hfill
  \begin{subfigure}[t]{0.45\textwidth}
    \includegraphics[width=\textwidth]{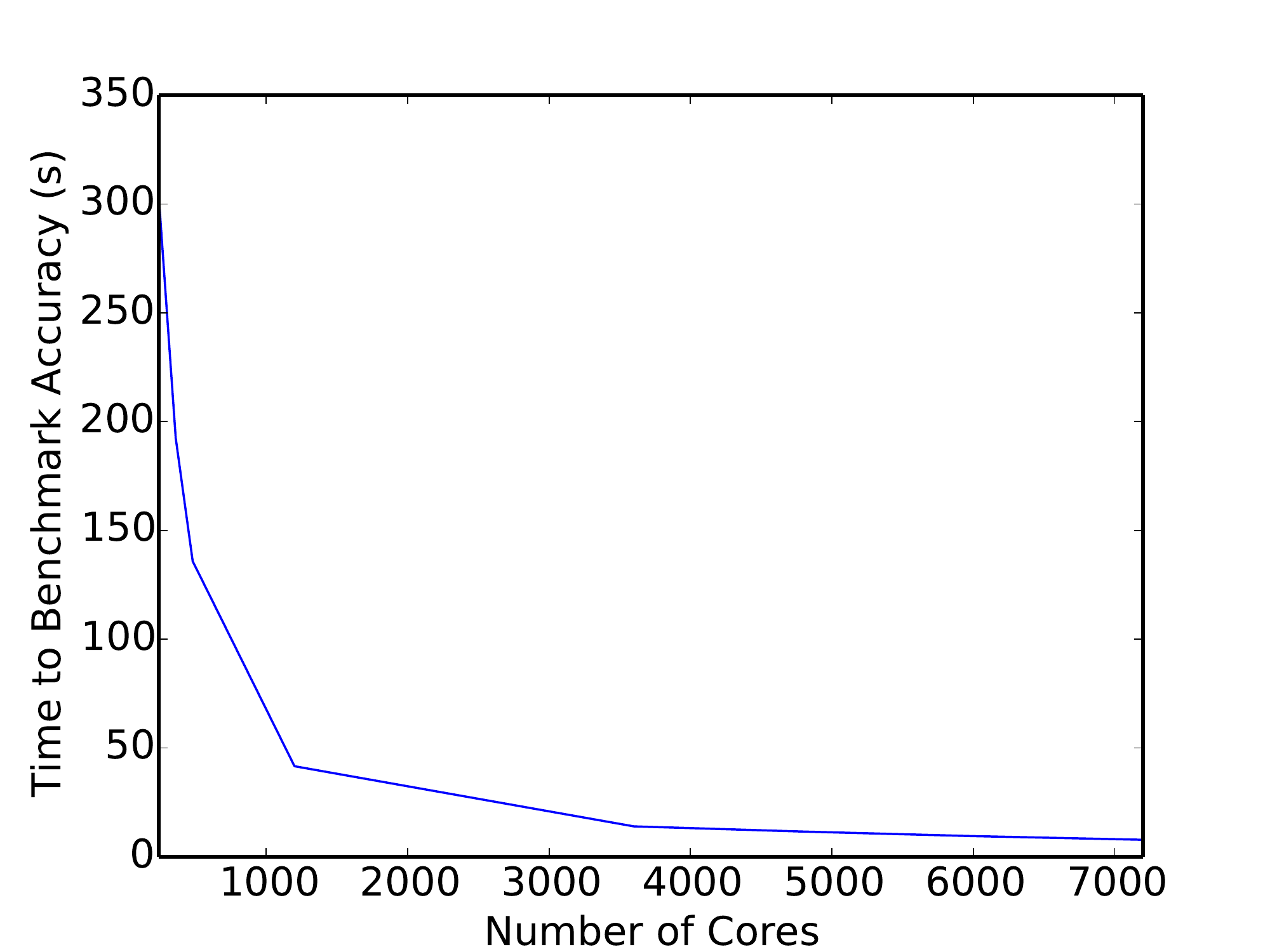}
    \caption{\textbf{Time required for ADMM to reach 64\% test accuracy when
      parallelized over varying levels of cores}.  L-BFGS on a GPU required 181
    seconds, and conjugate gradients required 44 minutes.  SGD never reached
    64\% accuracy.}
    \label{sfig:higgsScaling}
  \end{subfigure}
  \hfill
  \begin{subfigure}[t]{0.45\textwidth}
    \includegraphics[width=\textwidth]{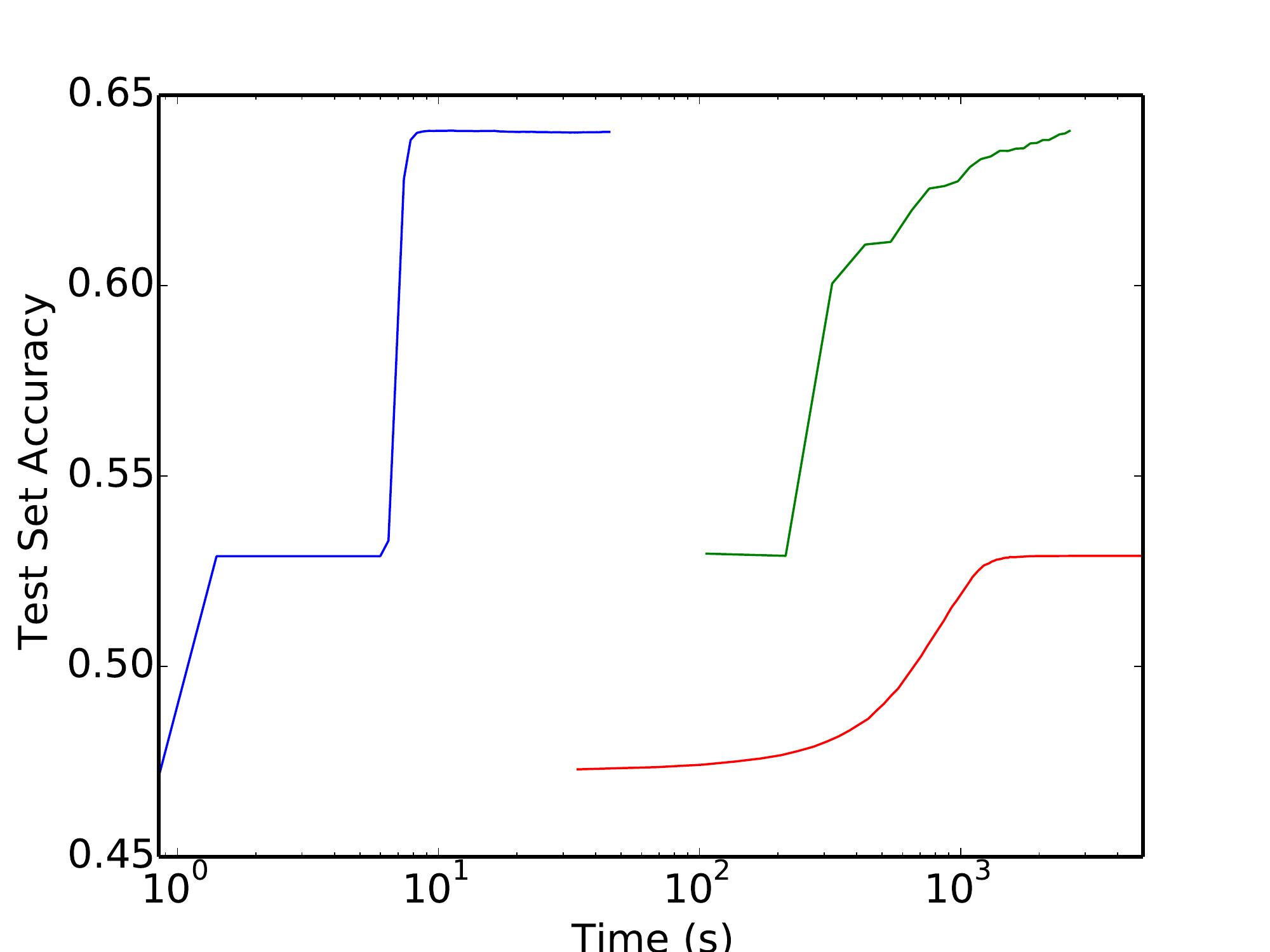}
    \caption{\textbf{Test set predictive accuracy as a function of time} for
    ADMM on 7200 cores (blue), conjugate gradients (green), and SGD (red).
  Note the x-axis is scaled logarithmically.}
    \label{sfig:higgsConverge}
  \end{subfigure}
  \hfill
  \vspace{-1mm}
  \caption{Higgs (subsection \ref{ssec:higgs})}
  \label{fig:converge}
    \vspace{-3mm}
\end{figure*}

In this section, we present experimental results that compare the performance
of the ADMM method to other approaches, including SGD, conjugate gradients,
and L-BFGS on benchmark classification tasks.  Comparisons are made across
multiple axes.  First, we illustrate the scaling of the approach, by varying
the number of cores available and clocking the compute time necessary to meet
an accuracy threshold on the test set of the problem.  Second, we show test
set classification accuracy as a function of time to compare the rate of
convergence of the optimization methods.  Finally, we show these comparisons
on two different data sets, one small and relatively easy, and one large and
difficult.

The new ADMM approach was implemented in Python on a Cray XC30 supercomputer
with Ivy Bridge processors, and communication between cores performed via MPI.
SGD, conjugate gradients, and L-BFGS are run as implemented in the Torch optim
package on NVIDIA Tesla K40 GPUs.  These methods underwent a thorough
hyperparameter grid search to identify the algorithm parameters that produced
the best results.  In all cases, timings indicate only the time spent
optimizing, excluding time spent loading data and setting up the network.

Experiments were run on two datasets.  The first is a subset of the Street
View House Numbers (SVHN) dataset~\cite{netzer2011reading}.  Neural nets were
constructed to classify pictures of 0s from 2s using histogram of gradient
(HOG) features of the original dataset.  Using the ``extra" dataset to train,
this meant 120,290 training datapoints of 648 features each.  The testing set
contained 5,893 data points.

The second dataset is the far more difficult Higgs
dataset~\cite{baldi2014searching}, consisting of a training set of 10,500,000
datapoints of 28 features each, with each datapoint labelled as either a
signal process producing a Higgs boson or a background process which does not.
The testing set consists of 500,000 datapoints.

\subsection{SVHN}
\label{ssec:svhn}

First, we focus on the problem posed by the SVHN dataset.  For this dataset,
we optimized a net with two hidden layers of 100 and 50 nodes and ReLU
activation functions.  This is an easy problem (test accuracy rises quickly) that does not require a large volume of data
and is easily handled by gradient-based methods on a GPU.  However, Figure
\ref{sfig:svScaling} demonstrates that ADMM exhibits linear scaling with
cores.  Even though the implementations of the gradient-based methods enjoy
communication via shared memory on the GPU while ADMM required CPU-to-CPU
communication, strong scaling allows ADMM on CPU cores to compete with the
gradient-based methods on a GPU.

This is illustrated clearly in Figure \ref{sfig:svConverge}, which shows each
method's performance on the test set as a function of time.  With 1,024
compute cores, on an average of 10 runs, ADMM was able to meet the 95\% test
set accuracy threshold in 13.3 seconds. After an extensive hyperparameter
search to find the settings which resulted in the fastest convergence, SGD
converged on average in 28.3 seconds, L-BFGS in 3.3 seconds, and conjugate
gradients in 10.1 seconds.  Though the small dataset kept ADMM from taking
full advantage of its scalability, it was nonetheless sufficient to allow it
to be competitive with GPU implementations.

\subsection{Higgs}
\label{ssec:higgs}

For the much larger and more difficult Higgs dataset, we optimized a simple
network with ReLU activation functions and a hidden layer of 300 nodes, as
suggested in \cite{baldi2014searching}.  The graph illustrates the amount of
time required to optimize the network to a test set prediction accuracy of
64\%; this parameter was chosen as all batch methods being tested reliably hit
this accuracy benchmark over numerous trials.  As is clear from Figure
\ref{sfig:higgsScaling}, parallelizing over additional cores decreases the
time required dramatically, and again exhibits linear scaling.

In this much larger problem, the advantageous scaling allowed ADMM to reach
the 64\% benchmark much faster than the other approaches.  Figure
\ref{sfig:higgsConverge} illustrates this clearly, with ADMM running on 7200
cores reaching this benchmark in 7.8 seconds.  In comparison,
L-BFGS required 181 seconds, and conjugate gradients required 44
minutes.\footnote{It is worth noting that though L-BFGS required substantially
more time to reach 64\% than did ADMM, it was the only method to produce a
superior classifier, doing as well as 75\% accuracy on the test set.}  In
seven hours of training, SGD never reached 64\% accuracy on the test set.
These results suggest that, for large and difficult problems, the strong
linear scaling of ADMM enables it to leverage large numbers of cores to
(dramatically) out-perform GPU implementations.   

\section{Discussion \& Conclusion}
\label{sec:discussion}
We present a method for training neural networks without using gradient steps.
In addition to avoiding many difficulties of gradient methods (like saturation
and choice of learning rates), performance of the proposed method scales
linearly up to thousands of cores.  This strong scaling enables the
proposed approach to out-perform other methods on problems involving extremely large datasets.

\subsection{Looking forward}
The experiments shown here represent a fairly narrow range of classification
problems and are not meant to demonstrate the absolute superiority of ADMM as a
training method.   Rather, this study is meant to be a proof of concept
demonstrating that the caveats of gradient-based methods can be avoided using
alternative minimization schemes.  Future work will explore the behavior of
alternating direction methods in broader contexts.  

We are particularly interested in focusing future work on recurrent nets and
convolutional nets.  Recurrent nets, which complicate standard
gradient methods \cite{jaeger2002tutorial,lukovsevivcius2012practical}, pose
no difficulty for ADMM schemes whatsoever because they decouple
layers using auxiliary variables.  Convolutional
networks are also of interest because ADMM can, in principle, handle them very
efficiently.  When the linear operators $\{W_l\}$ represent convolutions
rather than dense weight matrices, the least squares problems that arise in
the updates for $\{W_l\}$ and $\{a_l\}$ can be solved efficiently using fast
Fourier transforms.

Finally, there are avenues to explore to potentially improve
convergence speed.  These include adding momentum terms
to the weight updates and studying different initialization schemes, both
of which are known to be important for gradient-based schemes \cite{sutskever2013importance}.

\section*{Acknowledgements}
This work was supported by the National Science Foundation (\#1535902), the
Office of Naval Research (\#N00014-15-1-2676 and \#N0001415WX01341), and the
DoD High Performance Computing Center.

\bibliography{admm_nets}
\bibliographystyle{icml2016}

\end{document}